\documentclass[conference]{IEEEtran}
\usepackage[utf8]{inputenc}
\usepackage{graphicx}
\usepackage{amsmath,amsfonts,amssymb}
\usepackage{array}
\usepackage{multirow}
\usepackage{enumitem}

\newcolumntype{C}[1]{>{\centering\arraybackslash}p{#1}}

\title{Organized Event Participant Prediction Enhanced by Social Media Retweeting Data}
\author{
\IEEEauthorblockN{Yihong Zhang and Takahiro Hara}
\IEEEauthorblockA{ Graduate School of Information Science and Technology, 
Osaka University, Osaka, Japan\\
\{zhang.yihong, hara\}@ist.osaka-u.ac.jp
}
}

\begin{document}

\maketitle
\begin{abstract}
Nowadays, many platforms on the Web offer organized events, allowing users to be organizers or participants. For such platforms, it is beneficial to predict potential event participants. Existing work on this problem tends to borrow recommendation techniques. However, compared to e-commerce items and purchases, events and participation are usually of a much smaller frequency, and the data may be insufficient to learn an accurate model. In this paper, we propose to utilize social media retweeting activity data to enhance the learning of event participant prediction models. We create a joint knowledge graph to bridge the social media and the target domain, assuming that event descriptions and tweets are written in the same language. Furthermore, we propose a learning model that utilizes retweeting information for the target domain prediction more effectively. We conduct comprehensive experiments in two scenarios with real-world data. In each scenario, we set up training data of different sizes, as well as warm and cold test cases. The evaluation results show that our approach consistently outperforms several baseline models, especially with the warm test cases, and when target domain data is limited.

\end{abstract}

\section{Introduction}
Many digital platforms now are offering organized events through the Internet, where users can be organizers or participants. For example, the platform Meetup\footnote{https://www.meetup.com/} allow people to organize offline gatherings through online registration. And there are flash sales platforms such as Gilt\footnote{https://www.gilt.com/} that offer limited-time product discounts. Moreover, retweeting viral messages of the moment on social media platforms such as Twitter\footnote{https://www.twitter.com} can be also considered a type of event. Effectively predicting event participants can provide many benefits to event organizers and participants. For example, organizers can send out invitations more effectively \cite{yu2015should}, while potential participants can receive better recommendations \cite{qiao2014event}. Some previous researches have found that the problem of event participant prediction can be solved with recommendation techniques, such as matrix factorization \cite{jiang2019should}. Indeed if one considers events as items, and participation as users, then recommending events to users can be performed similarly as recommending products to users with an e-commerce recommender system \cite{sarwar2001item}. Unlike a product-based e-commerce platform, though, which has thousands of items, each purchased by thousands of users, events are organized and participated with much smaller frequency. Therefore, one problem with many event-based platforms is that they have not collected enough data to effectively learn a model of user preferences.

On the other hand, social media platforms such as Twitter nowadays are generating huge amounts of data that are accessible publicly \cite{who-says-what-twitter}. A particular activity, that is \emph{retweeting}, in which social media users repeat a popular tweet, can be seen as a type of event participant \cite{gao2015modeling}. We argue that event-based platforms can use data of such activity to support their own prediction models even though some restrictions are required. For example, due to privacy concerns, it is assumed that users in the target domain will not offer their social media account information. This condition invalidates many cross-domain recommendation solutions that rely on linked accounts \cite{deng2015twitter,zhao2015connecting,hu2018conet}. Nevertheless, even if the users are not linked to social media accounts, we can still have some useful information from social media. For examples, the interaction data that consists of user retweeting records of past tweets, and the tweet texts that are written in the same natural language. Retweeting data are useful for event participant prediction because the act of retweeting generally reveals a user's preference towards what is described in the tweet text \cite{gruhl2005predictive,atouati2020negative}. 

In this paper, we propose a method to utilize social media retweeting data during the learning of an event participant prediction model of a target domain, which has limited training data. As mentioned, we do not assume there are linkable users across social media and the target domain. Instead, we only assume that the event descriptions in the target domains are written in the same language as the social media tweets. This will become our basis for linking two domains. We generate a joint graph using data from two domains, and learn cross-domain users embeddings in the same embedding space. In this way, we can increase training data by adding social media retweeting data, and train more accurate models. To the best of our knowledge, this is the first work to use social media retweeting to enhance event participant prediction.

\section{Related Work}
We follow the recent research trend of event participant prediction, which is identified as an important problem in event-based social network (EBSN). Previously, Liu et al. studied the participant prediction problem in the context of EBSN \cite{liu2012event}. Their technique relied on the topological structure of the EBSN and early responded users. Similarly, Zhang et al. \cite{zhang2015will} proposed to engineer some user features and then apply machine learning such as logistic regression, decision tree, and support vector machines. Additionally, Du et al. considered the event descriptions, which were overlooked in previous works \cite{du2014predicting}. As the matrix factorization became a standard method in recommendation systems \cite{chin2015learning,xiao2018time}, later works also attempted to use this method in participant prediction. For example, Jiang and Li proposed to solve the problem by engineering user features and applying feature-based matrix factorization \cite{jiang2019should}. In this paper, we propose a prediction framework build on top of a deep neural network model of matrix factorization \cite{he2017neural}. In contrast to existing works, our framework is designed to use social media retweeting data to enhance the recommendation performance in the target domain.

Our inspiration comes from various works that use a support domain to help solve computation problems in a target domain. Especially, social media has been used in various works as the support domain. For example, Wei et al. have found that Twitter volume spikes could be used to predict stock options pricing \cite{wei2016twitter}. Asur and Huberman studied if social media chatter can be used to predict movie sales \cite{predicting-movie-revenues-twitter}. Pai and Liu proposed to use tweets and stock market values to predict vehicle sales \cite{pai2018predicting}. Broniatowski et al. made an attempt to track influenza with tweets \cite{broniatowski2015using}. They combined Google Flue Trend with tweets to track municipal-level influenza. These works, however, only used high-level features of social media, such as message counts or aggregated sentiment scores. In this work, we consider a more general setting of using retweeting as a supporting source to help participation prediction in the target domain, and users and events are transformed into embeddings for a wider applicability.

\section{Problem Formulation}
We formulate the problem of event participant prediction leveraging social media retweeting data as the following. In the target domain, we have a set of event data $E^T$, and for each event $e \in E^T$, there is a number of participants $p(e)=\{u^T_1,...,u^T_n\}$, with $u_i \in U^T$. In the social media retweeting data, we have a set of tweets $E^S$, for $e \in E^S$, we have retweeters $p(e)=\{u^S_1,...,u^S_m\}$, with $u_i \in U^S$. Normally we have fewer event data in the target domain than in the retweeting data, so $|E^S| > |E^T|$. We assume no identifiable common users across two domains, so $U^T \cap U^S = \emptyset$. An event in the target domain is described using the same language as the tweets. Let $d(e) = \{w_i,...,w_l\}$ be the words in the description of event $e$. If $V^S$ and $V^T$ are the description vocabularies in the tweets and the target domain, then $V^S \cap V^T \neq \emptyset$.

We can represent event descriptions and users as vector-form embeddings. Since the event descriptions in the target domain and the tweet texts are written in the same language, their embeddings can also be obtained from the same embedding space. We denote $r(e)$ as the function to obtain embeddings for event $e$ for both the target domain events and tweets. In the target domain, we have base user embeddings $l^B(u)$ available through the information provided by the platform user.

\section{Entity-connected Graph for Learning Joint User Embedding}
There exists a number of established techniques that learn embeddings from graphs \cite{bordes2013translating}. Our method is to learn a joint embedding function for both target domain and social media users by deploying such techniques, after creating a graph that connects them. Based on the participation data, we can create four kinds of relations in the graph, namely, participation relation, co-occurrence relation, same-word relation, and word-topic relation. 


The participation relation comes from the interaction data, and is set between users and words of the event. Suppose user $u$ participates in event $e$. Then we create $rel(u, w) = participation$, for each word $w$ in $d(e)$.

The co-occurrence relation comes from the occurrence of words in the event description. We use \emph{mutual information} \cite{peng2005feature} to represent the co-occurrence behavior. Specifically, we have $mi(w_1, w2) = log(\frac{N(w_1, w_2)|E|}{N(w_1)N(w_2)}$, where $N(w_1, w_2)$ is the frequency of co-occurrence of words $w_1$ and $w_2$, $|E|$ is the total number of events, and $N(w)$ is the frequency of occurrence of a single word $w$. We use a threshold $\phi$ to determine the co-occurrence relation, such that if $mi(w_1, w_2) > \phi$, we create $rel(w_1, w_2) = co\_occurrence$.

Two kinds of relations mentioned above are created within a single domain. We now connect the graph of two domains using the same-word relation. We create $rel(w^T, w^S) = same\_word$, if a word in the target domain and a word in the retweeting data are the same word. In this way, two separate graphs for two domains are connected through entities in the event descriptions. Once we have the joint graph, we can use established graph embedding learning techniques to learn user embeddings. In our experiment, we use TransE \cite{bordes2013translating} as the embedding learning technique.

\section{Event Participant Prediction Leveraging Joint User Embeddings}
We have shown how to obtain joint user embeddings for two event domains. Now we need a method to use them for the problem we aim to solve, that is event participant prediction. In this section, we will discuss first how event participant prediction can be solved in a single domain. Then we will present our framework that leverages joint user embeddings to solve the problem.

\subsection{Single Domain Prediction}
We find that the event participant prediction can be solved by recommendation techniques. Similar to the user and item interaction in a recommendation problem, event participation can be also treated as the interaction between users and events. After considering several options, we choose the state-of-the-art cold-start recommendation model proposed by Wang et al. \cite{wang2020dnn}. It is a generalization of a neural matrix factorization (NeuMF) model \cite{he2017neural} which originally used one-hot representation for users and items.

We aim to use the model to learn the following function:
\begin{equation}
    \hat{y}_{ue} = f(l(u), r(e))
\end{equation}
where $l(u)$ and $r(e)$ are the learned embeddings for user $u$ and event $e$. NeuMF ensembles two recommendation models, called generalized matrix factorization (GMF) and multi-layer perceptron (MLP). Specifically, it makes prediction:
\begin{equation}
    \label{eqn:neumf-concat}
    f(l(u), r(e)) = \sigma \left[GMF(l(u), r(e)) \bigoplus MLP(l(u), r(e)) \right]
\end{equation}
where $\sigma$ is a linear mapping function, and $\bigoplus$ is a concatenation operation.

Since the dataset usually contains only observed interactions, i.e., user purchase records of items, when training the model, it is necessary to bring up some negative samples, for example, by randomly choosing some user-item (user-event) pairs that have no interaction. The loss function for participant prediction is defined as the following:
\begin{equation}
\mathcal{L}_{PartP} = \sum_{(u, e) \in \mathcal{Y} \cup \mathcal{Y}^-}
y_{ue} \log \hat{y}_{ue} + (1-y_{ue})\log (1-\hat{y}_{ue}),
\end{equation}
where $y_{ue} = 1$ if user $u$ participated in event $e$, and 0 otherwise. $\mathcal{Y}$ denotes observed interactions and $\mathcal{Y}^-$ denotes negative samples.

\subsection{Leveraging Joint User Embeddings}
We have acquired in the previous section joint user embeddings,  $l^J(u)$, from the entity-connected graph. Note that we can apply the same graph technique to learn embeddings in single domains as well, denoted as $l^S(u)$ and $l^T(u)$ respectively for the retweeting data and target domain. From problem formulation, we also have base user embedding for the target domain $l^B(u)$. A problem is that the graph embeddings $l^J(u)$ and $l^T(u)$ are only available for a small number of target domain users, because they are learned from limited participation data. When we predict participants in future events, we need to consider the majority of users who have not participated in past events. These users have base embeddings $l^B(u)$ but not graph embeddings $l^J(u)$ and $l^T(u)$.

We need to map base embedding $l^B(u)$ to the embedding space of $l^J(u)$ when making the prediction. As some previous works proposed, this can be done through linear latent space mapping \cite{man2017cross}. Essentially it is to find a transfer matrix $M$ so that $M \times U^s_i$ approximates $U^t_i$, and $M$ can be found by solving the following optimization problem
\begin{equation}
\min_M \sum_{u_i \in \mathbf{U}} \mathcal{L}(M \times U^s_i, U^t_i) + \Omega(M),
\end{equation}
where $\mathcal{L}(.,.)$ is the loss function and $\Omega(M)$ is the regularization. After obtaining $M$ from users who have both base embeddings and graph embeddings, we can map the base user embedding to graph user embedding $l^{J'}(u) = M \times l^B(u)$ for those users who have no graph embedding.

An alternative solution would be using the base user embedding as the input for training the model. This would then require us to map graph user embedding to target domain base user embedding. Unlike mapping base embedding to graph embedding, where some target domain users have both embeddings, we do not have social media users with base embeddings. So the mapping requires a different technique. We solve it by finding the most similar target domain users for a social media user, and using their embeddings as the social media user base embedding. More specifically, we pick $k$ most similar target domain users according to the graph embedding, and take the average of their base embedding:
\begin{equation}
    l^{B'}(u) = \frac{1}{K} \sum_{u_i \in U^K} l^B(u_i)
\end{equation}
where $U^K$ is top-k target domain users most similar to the social media user $u$ according to their graph embeddings.

\subsection{Base and Graph Fusion}
We have shown two ways to create joint training data by mapping graph embeddings to base embeddings, and by mapping base embeddings to graph embeddings. Both embedding spaces have their advantages. The graph embeddings are taken from the interaction data, thus contain information useful for predicting participation. The base embedding contains user context obtained from the target domain, which can supply extra information. While it is possible to use the two types of embeddings separately, we would like to propose a fusion unit that leverages the advantages of both embedding spaces. We call the method base and graph embedding fusion (BGF).


After obtaining training data for two types of embeddings, we train two prediction models separately for them using the NeuMF model. The input event embeddings $r(e)$ are the same for both models. The input user embeddings are selected depending on whether the user has a graph embedding available or not. More specifically, for graph embedding space, the input $l(u)$ is set to $l^J(u)$ if user $u$ has graph embedding, and otherwise it is set to the mapped embedding $l^{J'}(u)$. Similarly we do for the base embedding space, and select either $l^B(u)$ or $l^{B'}(u)$ depending on the availability. Then, instead of output predictions, we take the concatenation layers of two NeuMF models, produced by the concatenation in Equation (\ref{eqn:neumf-concat}) and concatenate them together. The prediction is made on the output of this large concatenation layer.

Following a recent trend in deep learning research, we use an attention module \cite{vaswani2017attention} to further refine the output of the model. An attention module is generally effective when we need to choose more important information from inputs. Since after running two prediction models, we have a large number of information units, it is suitable to apply the attention module.

The idea of attention is to use a vector query to assign weights to a matrix so the more important factors can be emphasized. The query is compared with keys, a reference source, to produce a set of weights, which is then used to combine the candidate embeddings. For the current scenario, we use the concatenated output of NeuMF as the key and the event embedding as the query. The output of the attention module is a context vector $c_i$ for event $i$
\begin{equation}
    \mathbf{c}_i = \sum_j a_{ij} s_j
\end{equation}
where $a_{ij}$ is attention weights, and $s_j$ is the key. We transform the concatenated output of NeuMF into a matrix with the same number of columns as the query dimensions, and use it as the key $s_j$. The attention weights can be obtained using the $general$ attention score \cite{luong2015effective}.

We insert the attention module after the output of two prediction models and use the event embedding as the query to select the more important information. Empirically, we do find adding the attention module improves overall prediction accuracy.

We note that BGF can be used with a single domain. We can construct the graph for a single domain without the bridging relations, i.e., only keeping the word co-occurrence and the user participation relations. Using the described above, we can have two sets of embedding generated, from the base embedding and from the graph, and on them the BGF unit can be applied. In the empirical study to be presented later, the single domain BGF is shown to have achieved relatively high prediction accuracy.

\subsection{Leveraging Cross-domain Learning}
We have integrated social media retweeting into the event participation data of a target domain using the method described above. Now we can simply combine the retweeting data with the event participant data, treating them as a single dataset. However, there are better ways to train the model across domains, as proposed by recent studies in transfer learning. Here we will introduce a transfer learning technique that can be used to further improve our method.


The technique is called knowledge distillation (KD) \cite{aguilar2020knowledge}. It has been shown that, when model learning is shifted from one task to another task, this technique can be used to distill knowledge learned in the previous task. The distilled knowledge becomes accessible through the KL-divergence, a measure of the difference between prediction results using the new model and the old model. Specifically, we set up a loss through KL-Divergence:
\begin{equation}
    \mathcal{L}_{KD} = D_{KL}(\hat{Y}_{new} || \hat{Y}_{old})
\end{equation}
where $\hat{Y}_{new}$ and $\hat{Y}_{old}$ are predictions made with the model learned in the new domain and the old domain, respectively, and $D_{KL}$ is the point-wise KL-Divergence.

We first train the model using the retweeting data, and then shift to the target domain participation data. The single domain loss and the KD loss can be counted together in the cross-domain model learning, as
\begin{equation}
    \mathcal{L}_{CD} = \mathcal{L}_{PartP} + \mathcal{L}_{KD}.
\end{equation}

\section{Experimental Evaluation}
To verify the effectiveness of our approach, we perform experiments with a public event dataset, taken from event platform Meetup. On Meetup, events are explicitly defined by organizers, and users register for participation. We use Twitter as the supporting social media source. In this section, we will discuss the dataset preparation and the experiment setup, before presenting the evaluation results.

\subsection{Dataset Collection}
We use a publicly available dataset\footnote{https://ieee-dataport.org/documents/meetup-dataset}. The dataset was collected for the purpose of analyzing information flow in event-based social networks \cite{liu2012event}. On Meetup, users can participate in social events, which are only active in a limited period, or they can join groups, which do not have time restriction. Events and users are also associated with tags, which are associated with descriptive English keywords. Popular event examples are language study gatherings, jogging and hiking sessions, and wine tasting workshops. The dataset contains relations between several thousands of users, events, and groups. Our interest is mostly in the user-event relation.

We prepare a corresponding Twitter retweet dataset. We monitor Twitter for tweets authored by users with the keyword ``she/her'' and ``he/him'' in their profile description, which results in more than two million tweets. While these tweets covers many topics, they are more or less gender-aware given the author profiles. We construct retweet clusters from these retweets, and obtain several thousands of retweet clusters, each retweeted at least ten times by users in the dataset.

Since our objective is to investigate the effect of adding retweets when the target domain has limited data, we generate datasets of different sizes. Specifically, we select three sizes of datasets, containing 100, 200, and 500 events. To balance the retweets with event data, we use the same number of tweets as the events. The events are randomly selected, and the tweets are also randomly selected with the restriction that their texts have common words with the event descriptions. The number of events, users, participation, tweets, Twitter users, and retweets are shown in Table \ref{tab:data}.

\begin{table}[ht]
    \centering
    \caption{Experimental dataset statistics}
    \label{tab:data}
    \begin{tabular}{c | c c c c c }
Events&Users&Participation&Tweets&SM Users&Retweets\\ \hline
100&448&1,792&100&1,042&5,960\\
200&898&3,592&200&2,255&12,612\\
500&2,460&9,840&500&6,599&36,236\\
    \end{tabular}
\end{table}

We use pre-trained embeddings to represent event descriptions and tweets of the same language. Specifically, we use the Spacy\footnote{https://spacy.io/} package, which provides word embeddings trained on Web data, and a pipeline to transform sentences into embeddings. For our approach, we also need to provide base user embeddings. For the Meetup dataset, the users are associated with tags, which are associated with text keywords. We again use Spacy to transform user tags to embeddings, and use them as the base user embedding.

\subsection{Experiment Setup}
We set up two test cases, based on whether or not test data contain events in the training data. In the case where test data contain events in the training data, which is called \emph{warm test}, we randomly pick up one user from each event, adding it to the test data and removing it from the training data. In the case where test data contain no event in the training data, which is called \emph{cold test}, we use all data shown in Table \ref{tab:data} as the training data, and use additional 1,000 events as the test data.

We create the training dataset by random negative sampling. For every interaction entry $(u, e)$ in the training dataset, which is labeled as positive, we randomly pick four users who have not participated in the event, and label the pairs as negative. The testing is done by event. For each event $e$ in the test dataset, we label all users who participated in the event $U^+$ as positive. Then, for the purpose of consistent measurement, we pick $n - |U^+|$ users, labeled as negative, so that the total candidate is $n$, which is set to 100. For the warm test, $|U^+|=1$, while for the cold test, $|U^+|$ varies from event to event.

We predict the user preference score for all the $n$ users, rank them by the score, and measure the prediction accuracy based on top $k$ users in the rank. We measure $Recall@10$ and $Precision@5$.

We compare our method with three baselines in the existing literature, in addition to variations of our own approaches. The baselines include:
\begin{itemize}[leftmargin=*]
    \item base, which runs the recommendation model on target domain base embeddings.
    \item BGF, the base and graph fusion model we introduced. In this variation, it is used with only the target domain data.
    \item MIX, a variation of our approach without the knowledge distillation component. Instead, it mixes target domain participation data and the retweets as a single training dataset.
    \item BPRMF \cite{rendle2009bpr}, a single domain matrix factorization-based recommendation model, known for its effectiveness in implicit recommendation.
    \item CKE \cite{zhang2016collaborative}, a knowledge graph-based recommendation model. It can be used for cross-domain prediction if the supporting domain is transformed into a knowledge graph.
    \item KGAT \cite{wang2019kgat}, a state-of-the-art knowledge graph-based recommendation model. It can be used for cross-domain prediction like CKE. However, it does not deal with cold items so we skip it for the cold test.
\end{itemize}

We implement our approach and all baselines in Python and Tensorflow. We set 200 as the latent factor embedding size where it is needed.

\subsection{Evaluation Results and Discussions}
The experimental results are shown in Tables \ref{tab:acc-meetup}, respectively. Single-domain methods are indicated by (SD) and cross-domain methods are indicated by (CD). The best results in each test are highlighted in bold font.
\begin{table}[ht]
    \centering
    \caption{Prediction accuracy for the Meetup dataset}
    \label{tab:acc-meetup}
    \begin{tabular}{l|c c |c c |c c}
\hline
&\multicolumn{2}{c|}{Meetup 100}&\multicolumn{2}{c|}{Meetup 200}&\multicolumn{2}{c}{Meetup 500}\\ \hline
&R@10&P@5&R@10&P@5&R@10&P@5\\ \hline
\multicolumn{7}{l}{warm test}\\ \hline
(SD) base &0.192 &0.027 &0.196 &0.030 &0.190 &0.016 \\
(SD) BPRMF &0.135 &0.019 &0.098 &0.007 &0.061 &0.006 \\
(SD) BGF &0.385 &0.050 &0.589 &0.080 &0.887 &0.154 \\
(CD) CKE &0.135 &0.012 &0.125 &0.011 &0.103 &0.010 \\
(CD) KGAT &0.250 &0.042 &0.116 &0.020 &0.090 &0.007 \\
(CD) MIX &0.154 &0.023 &0.196 &0.023 &0.190 &0.021 \\
(CD) proposed &\textbf{0.404} &\textbf{0.073} &\textbf{0.688} &\textbf{0.105} &\textbf{0.955} &\textbf{0.172} \\ \hline
\multicolumn{7}{l}{cold test}\\ \hline
(SD) base&0.106 &0.054 &0.115 &0.055 &\textbf{0.135} &\textbf{0.067} \\
(SD) BPRMF &0.097 &0.049 &0.094 &0.050 &0.094 &0.044 \\
(SD) BGF &0.121 &0.055 &0.098 &0.042 &0.065 &0.023 \\
(CD) CKE &0.098 &0.048 &0.100 &0.051 &0.089 &0.043 \\
(CD) MIX &0.106 &0.049 &0.120 &\textbf{0.059} &0.065 &0.023 \\
(CD) proposed &\textbf{0.124} &\textbf{0.059} &\textbf{0.124} &0.057 &0.122 &0.050 \\ \hline
    \end{tabular}
\end{table}

First we look at the warm test. We can see that the proposed method has a clear advantage over other methods, achieving the best accuracy for both scenarios and for all training data sizes. Particularly, we see that it steadily outperforms MIX method, validating the effectiveness of knowledge distillation. The second best cross-domain method is KGAT, especially for smaller training data sizes. But its performance deteriorates as training data sizes increase. The best single-domain method, BGF, outperformed cross-domain methods like KGAT and CKE when training data size is large, showing the strength of fusing graph embedding and base embedding together. The proposed method, utilizes BGF and knowledge distillation, outperformed single domain BGF by up to 66\%.

Next we look at the cold test. We see that the result is more complex in the cold test. When the training data size is smaller, the proposed method generally shows some advantages. For example, when the training data size is 100, it achieves 2.4\% higher Recall@10 than BGF. When the training data size is 500, the base model achieves the best accuracy.

Comparing warm and cold tests, we see that cross-domain methods have an advantage for the former, but a disadvantage for the latter. The reason is that when we already have some participant data for an event, it is easier to use external knowledge to enhance the information. However, when there is no data for a new event, the useful information is mostly from the target domain itself, and retweeting data can only add limited useful information to the model, if not noises, especially when the target domain has sufficient training data.

\section{Conclusion}
In this paper, we propose to use social media retweeting data as an general enhancement for event participant prediction in a target domain. Our proposed solution involves a cross-domain knowledge graph, which assumes that event descriptions are written in the same language as social media tweets. We also present a learning method that utilizes joint user embedding from the knowledge graph and makes use of knowledge distillation. We test the method with real-world event participation data, comparing it with several baselines. And we show that our proposed method has clear advantage in terms of prediction accuracy, especially for the warm tests, where some participants of events are already known. For the cold test, we reach mixed results, with our method only superior in some training data sizes. 

\section*{Acknowledgement}
This research is partially supported by JST CREST Grant Number JPMJCR21F2.


\end{document}